\documentclass{article}
\usepackage{spconf,amsmath,graphicx}
\usepackage{algorithmic}
\usepackage{algorithm}
\usepackage{array}
\usepackage{bbold}
\usepackage{textcomp}
\usepackage{stfloats}
\usepackage{url}
\usepackage{verbatim}
\usepackage{graphicx,subcaption} 
\usepackage{cite}
\usepackage{lineno,hyperref}
\modulolinenumbers[5]
\usepackage{algorithmic}
\usepackage[dvipsnames]{xcolor}
\usepackage{mathtools}
\usepackage{enumerate}
\usepackage{booktabs}
\usepackage{lipsum}
\usepackage{color}
\newcommand{\mat}[1]{\boldsymbol{#1}}
\newcommand{\real}{\mathbb{R}}
\newcommand{\norm}[1]{\left\lVert#1\right\rVert}
\usepackage{graphicx,subcaption}

\title{Fast and Accurate Outlier-Aware LiDAR Super-Resolution for SLAM Applications}

\name{Christos Anagnostopoulos$^{1,2}$\qquad Alexandros Gkillas$^{1,3}$\qquad Nikos Piperigkos$^{1,3}$ \qquad Aris S. Lalos$^{1}$}

\address{$^{1}$ Industrial Systems Institute, Athena Research Center, Patras Science Park, Greece
\\
$^{2}$ Dpt. of Informatics \& Telecom., University of Ioannina, Arta, Greece
\\
$^{3}$ AviSense.AI, Patras Science Park, Greece
\thanks{This work has received funding from the European Union’s Horizon Europe research and innovation programme under the AutoTRUST project, “Autonomous Self-Adaptive Services for Transformational Personalized Inclusiveness and Resilience in Mobility” (Grant Agreement No. 101148123), and the GuardAI project, “Robust and Secure Edge AI Systems for Safety-Critical Applications” (Grant Agreement No. 101168067).}}

\begin{document}

%
\maketitle
\begingroup \renewcommand\thefootnote{} \footnotetext{\textcopyright{} 2025 IEEE. Personal use of this material is permitted. Permission from IEEE must be obtained for all other uses, in any current or future media, including reprinting/republishing this material for advertising or promotional purposes, creating new collective works, for resale or redistribution to servers or lists, or reuse of any copyrighted component of this work in other works. DOI: 10.1109/ICIP55913.2025.11084730 } \endgroup
\begin{abstract}
This work tackles the challenge of enhancing low-resolution LiDAR sensors for SLAM applications through a novel Deep Unrolling-based Super-Resolution (SR) model. We integrate an outlier removal module to ensure structural integrity while maintaining real-time performance. By leveraging a model-based optimization approach, our method efficiently reconstructs high-resolution point clouds while minimizing computational overhead. The proposed SR model is evaluated within a LiDAR SLAM framework, demonstrating significant improvements in pose estimation accuracy and efficiency compared to state-of-the-art SR methods. 
\end{abstract}
\section{Introduction}

\label{sec:intro}

LiDAR Simultaneous Localization and Mapping (SLAM) is essential for autonomous vehicles and robotics, enabling self-localization and environmental mapping. Compared to visual SLAM, LiDAR-based methods offer superior reliability in low-light or low-visibility conditions \cite{chghafCameraLiDARMultimodal2022}. By generating high-resolution 3D point clouds, LiDAR facilitates precise localization and navigation by aligning consecutive scans to minimize drift and improve trajectory estimation.
However, high-resolution LiDAR sensors are costly, making them impractical for many applications \cite{8936542}. While 64-channel LiDAR provides superior accuracy, lower-cost 16-channel sensors produce sparser point clouds, leading to reduced odometry accuracy, increased drift, and degraded SLAM performance \cite{10314010}. Additionally, higher resolution increases energy consumption, and costs. Balancing resolution, affordability, and odometry precision remains a key challenge in LiDAR-based SLAM.

Inevitably, research has focused on enhancing the resolution of low-cost LiDAR sensors to bridge the performance gap with high-resolution counterparts. This upsampling can be achieved through SR techniques, which refine low-resolution LiDAR data by reconstructing missing details and increasing spatial density \cite{lidar_method}. However, most SR approaches fail to account for their impact on SLAM accuracy. These methods typically process either raw 3D point clouds or 2D range images derived from LiDAR scans, yet they introduce several critical limitations that hinder their effectiveness in real-world SLAM applications \cite{lidar_method}, \cite{point_cloud}.
Deep learning-based SR algorithms outperform traditional interpolation methods but rely on complex architectures with numerous learnable parameters, requiring substantial high-quality training data. Their high computational complexity limits frame rates, making real-time processing impractical for SLAM applications. Additionally, these methods operate independently of SLAM optimization, leading to inconsistent high-resolution reconstructions and artifacts that propagate errors in the SLAM pipeline. Many SR techniques also introduce outliers, further degrading SLAM accuracy and requiring costly post-processing,  preventing real-time execution \cite{my_icip}.

Nevertheless, to achieve a resource-efficient architecture without the need of computationally intensive post-processing methods,  our approach draws its motivation from the model-based deep learning theory \cite{model_eldar}, leading a more efficient solution for the SR problem. To be more detailed, based on a novel optimization problem, we design a resource efficient model-based SR network utilizing the deep unrolling (DU) strategy \cite{model_eldar}. The resulting SR model can be integrated with the SLAM proces, resulting in an end-to-end real-time pipeline with improved accuracy. By leveraging the DU framework, we reformulate a novel  optimization problem into an efficient  deep learning architecture.
The main contributions of this work can be summarized as:
\begin{itemize}
\item \textbf{Model-based SR Network:} This work introduces a novel optimization problem for the LiDAR SR problem, formulated in the range-view domain. The cost function of the optimization framework consists of three key components: (1) a data-consistency module that enforces the mathematical relationship between the low-resolution and high-resolution range images, (2) an encoder-decoder-based regularizer that captures intrinsic properties of high-resolution range images, such as low-rank structures, and (3) a novel regularization term that simultaneously preserves structural consistency and removes outliers within the optimization process. By integrating this outlier removal process directly into the optimization framework, we eliminate the need for explicit post-processing while simultaneously enhancing the effectiveness of the encoder-decoder-based regularizer. The iterative solution of the proposed optimization problem is transformed into a model-based deep learning network using the deep unrolling strategy, resulting in a computationally efficient network, while effectively removes outliers and spurious points.
\item \textbf{End-to-End architecture}: The proposed computationally efficient SR model is combined with the SLAM to form the end-to-end framework.  Overall, the joint optimization and lightweight nature of the SR model allow the resulting end-to-end framework to achieve performance and computational efficiency. The proposed method was implemented in C++ and integrated into the LeGO-LOAM pipeline to further increase the computational efficiency in fps.
    \item \textbf{SOTA performance} We  evaluate our method  by testing it within a LiDAR-SLAM framework, demonstrating its superiority in pose estimation accuracy and efficiency compared to state-of-the-art SR methods.
\end{itemize}

\section{Related Work}

\subsection{LiDAR-based SLAM}
LiDAR SLAM systems can be categorized based on the type of sensors used. 3D LiDAR SLAM is more suitable for outdoor environments and is particularly well-suited for autonomous vehicle applications, as it captures data in three dimensions rather than being restricted to a single plane, as is the case with 2D LiDAR. They consist of a frontend and a backend. The frontend establishes data associations between consecutive point clouds and provides an initial pose estimate, while the backend minimizes accumulated errors using filter-based or graph-optimization techniques. A key challenge in the frontend is scan matching, which estimates the sensor’s relative pose by aligning point clouds. This can be approached via direct matching or feature matching.
Direct matching methods, such as ICP \cite{beslMethodRegistration3D1992} and its variants (Generalized-ICP \cite{segalGeneralizedICP2010}, or NICP \cite{serafinNICPDenseNormal2015}), iteratively minimize distances between corresponding points. By utilizing NDT \cite{biberNormalDistributionsTransform2003} or combining ICP with NDT \cite{hongProbabilisticNormalDistributions2017}, the requirement for explicit point correspondences can be eliminated, further improving efficiency. However, while scan-to-scan matching algorithms have advanced, their accuracy remains inherently limited. To improve pose estimation, recent LiDAR odometry techniques integrate both scan-to-scan and scan-to-map matching \cite{chenDirectLiDAROdometry2022}, \cite{deschaudIMLSSLAMScantomodelMatching2018}.
Feature matching methods extract  features to establish correspondences. LOAM \cite{zhangLOAMLidarOdometry2014} identifies linear and planar features based on curvature, while its extensions, such as RO-LOAM \cite{oelschROLOAM3DReference2022}, enhance trajectory estimation using mesh features. LeGO-LOAM \cite{shanLeGOLOAMLightweightGroundOptimized2018} improves feature extraction by segmenting points into ground and non-ground categories, utilizing planar and edge features to refine pose estimation.

\subsection{LiDAR-based Super-Resolution}

LiDAR Super-Resolution techniques can generally be classified into two main approaches. The first category focuses on performing SR directly on raw 3D LiDAR point clouds \cite{9388922, 9919373, 9875347}. However, these methods require substantial computational resources to identify neighboring point relationships and often necessitate additional processing steps, such as segmentation, due to the inherent sparsity of 3D point clouds \cite{ZHENG2024103783}. An alternative approach \cite{9811992, 10164213, SHAN2020103647} operates in the range image domain by projecting 3D point clouds onto 2D range images. This representation offers a more compact format, effectively mitigating data sparsity issues \cite{yang2024tuliptransformerupsamplinglidar}. However, these methods typically rely on deep learning architectures with a large number of parameters, such as U-Net-based networks \cite{SHAN2020103647} and vision transformers \cite{yang2024tuliptransformerupsamplinglidar}. Despite their potential, the computational demands of these models, combined with the need for post-processing to remove outliers, pose significant challenges for real-time implementation. The high complexity of inference, along with the additional processing overhead, often results in suboptimal frame rates, limiting their applicability in real-world SLAM.


\section{Proposed Methodology}
In this section, we formulate the basic SR LiDAR optimization problem by developing a model-based deep learning network to tackle the LiDAR SR problem, and integrating the SR method into an end-to-end LiDAR SLAM architecture.

\subsection{Optimization Problem}
To construct the proposed model-based deep learning architecture, we leverage the advantages of projection-based methodologies. Let $\mat{S_1}$ represent a high-resolution LiDAR sensor with $\mat{N_1}$ channels e.g., $64$, and $\mat{S_2}$ a low-resolution LiDAR sensor with $\mat{N_2}$ channels e.g., $16$, where $\mat{N_1}, \mat{N_2} \in \mathbb{Z}$ and $\mat{N_1} > \mat{N_2}$. We project the 3D high-resolution point cloud $\mat{P}$, obtained from $\mat{S_1}$, into a high-resolution range image $\mat{H} \in \mathbb{R}^{N_1\times K}$, where $K$ represents the horizontal resolution of the sensor. If $\mat{L} \in \mathbb{R}^{N_2\times K}$ denotes the range image produced by the low-resolution 16-channel LiDAR sensor, the transformation between the two range images i.e., the low-resolution and high-resolution image  can be described by the following mathematical relationship:
\begin{equation} \mat{L} = \mat{D} \mat{H} + \mat{E}, \label{eq:degr_model} \end{equation}
where  $\mat{D} \in \real^{N_2 \times N_1}$ denotes the downsampling operator that selects only the $N_2$ channels from the high resolution range image and $\mat{E}$ is a noise term.
Our goal is to upscale the quality of the resolution of the point cloud generated from the low resolution sensor.  To achieve this, we formulate an optimization problem based on the work presented in \cite{my_icip}:
\begin{equation}
\underset{\mat{H}}{\arg\min}\,\,\, \frac{1}{2} \norm{\mat{L} - \mat{D} \mat{H}}_F^2  + \mu \mathcal{J}(\mat{H}) 
\label{eq:main_problem_icip}
\end{equation}
where the first component  ensures consistency with the degradation model defined in \eqref{eq:degr_model}. The second component $\mathcal{J}(.)$ serves as a learnable regularizer, designed to capture the intrinsic features of the high-resolution range image  $\mat{H}$, and 
$\mu$ is the regularization parameter.

However, the above formulation of method \cite{my_icip} still faces challenges in ensuring real-time performance, as the learnable regularizer, which aims to capture the intrinsic properties of range images, relies on 2D convolutions. These convolutions, while effective in learning spatial patterns, do not inherently preserve the 3D geometric consistency of the original LiDAR point cloud. As a result, they often introduce outliers, particularly in regions with depth discontinuities, sharp edges, or occlusions. These artifacts often appear as spurious points in free-space regions or distortions along object boundaries, significantly affecting downstream SLAM tasks. To restore consistency, additional post-processing steps are often required to handle outliers. However, these steps increase computational overhead, restricting real-time performance and making the approach impractical for autonomous navigation.

To overcome this, we introduce a novel regularization term that simultaneously preserves structural consistency and removes outliers within the optimization process. By integrating the outlier removal process directly into the optimization framework, we eliminate the need for explicit post-processing while simultaneously enhancing the effectiveness of the learnable regularizer $\mathcal{J}(.)$. Since $\mathcal{J}(.)$ is trained jointly with the outlier removal module, it can learn to better capture the structural properties of the high-resolution range image while being aware of the effects of outlier suppression. This synergistic optimization leads to more accurate and stable SR outputs, ultimately improving LiDAR SLAM performance.  Thus, we propose a novel  optimization problem, defined as  
\begin{equation}
\underset{\mat{H}}{\arg\min}\,\,\, \frac{1}{2} \norm{\mat{L} - \mat{D} \mat{H}}_F^2  + \mu\mathcal{J}(\mat{H}) + \lambda \mathcal{Q}(\mat{H})
\label{eq:main_problem_seg}
\end{equation}
where the term $\mathcal{Q}(.)$ represents an outlier removal regularizer. This module guides the solution toward generating geometrically consistent high-resolution range images.

\subsection{Model-based deep learning SR} \label{DU}
In order to address the proposed optimization problem in (\ref{eq:main_problem_seg}), we utilize the Half quadratic splitting (HQS) methodology to decompose the initial problem into more manageable subproblems. Hence, the initial problem can be reformulated as:
\begin{align}
&\underset{\mat{H}}{\arg\min}\,\,\, \frac{1}{2} \norm{\mat{L} - \mat{D} \mat{H}}_F^2  + \mu\mathcal{J}(\mat{Z}) + \lambda \mathcal{Q}(\mat{Y}) \nonumber \\
&s.t. \,\,\,\, \mat{H} = \mat{Z}, \mat{H} = \mat{Y}
\label{eq:main_problem_seg_hqs}
\end{align}
where $\mat{Z} \in \real^{N_1\times K}$, $\mat{Y}\in \real^{N_1\times K}$ are auxiliary variables. 
\begin{figure*}
\centering
 \includegraphics[scale=0.4]{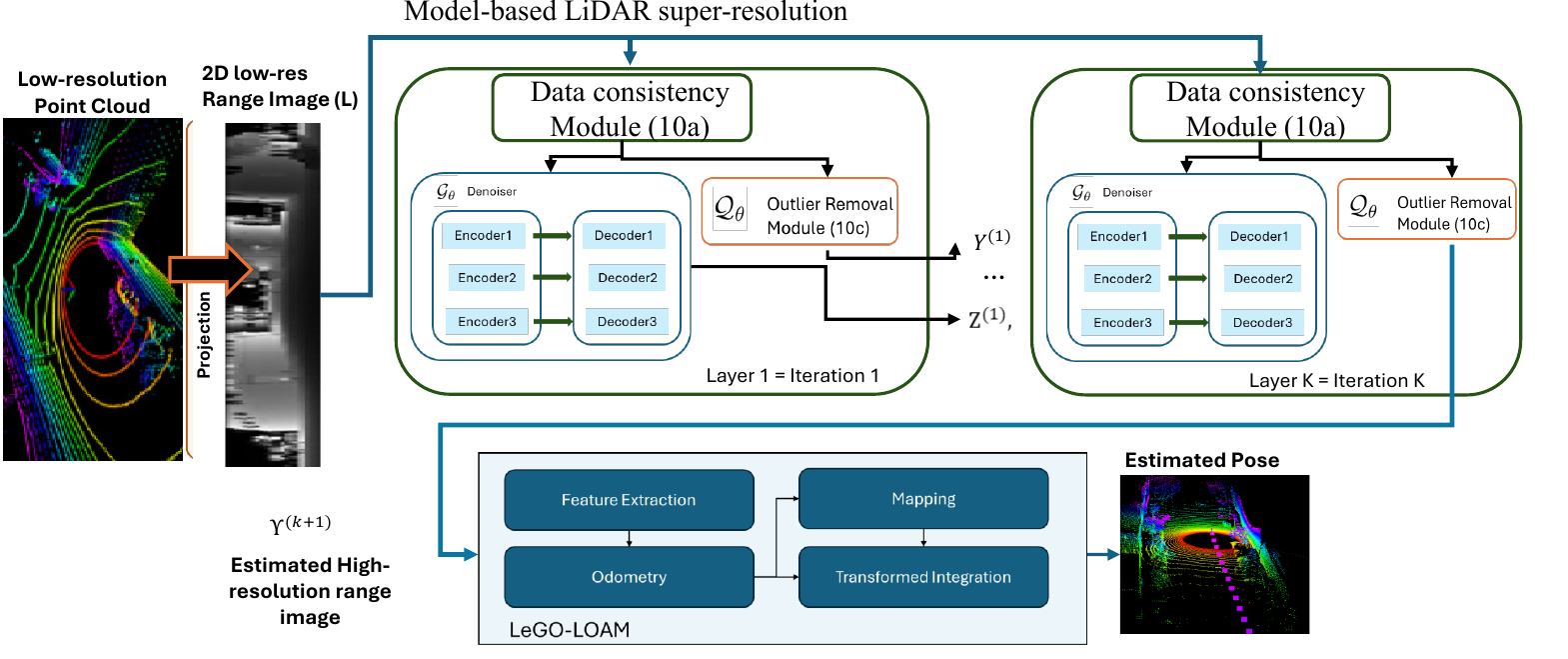}
  \caption{The proposed end-to-end Super-LiDAR-SLAM architecture. The pipeline starts with a low-resolution point cloud captured by a low resolution LiDAR sensor, which is projected into a 2D range image. The 2D range image is then upscaled using the SR module, that consists of two denoising process i.e., the autoencoder denoiser (10b) that aims to refines the output of the data-consistency module (10a) and the outlier removal module  to filter spurious points (10c). 
  The processed output is passed into the  LeGO-LOAM framework to estimate the final pose.
  }
  \label{fig:arghitecture}
\end{figure*}
The loss function that HQS method seeks to minimize is:
\begin{align}
    \mathcal{L} = \frac{1}{2}\norm{\mat{L} - \mat{D} \mat{H}}_F^2  &+ \mu\mathcal{J}(\mat{Z}) + \lambda \mathcal{Q}(\mat{Y}) \nonumber\\
    &+ \frac{b}{2}\norm{Z-H}_F^2+ \frac{b}{2}\norm{Y-H}_F^2
    \label{eq:Lagrangian}
\end{align}
where b denotes a penalty parameter. Based on equation (\ref{eq:Lagrangian}) a sequence of individual sub-problems emerges:
\begin{subequations}
\begin{align}
    \mat{H}^{(k+1)} = \underset{\mat{H}}{\arg\min}\,&\frac{1}{2} \norm{\mat{L} - \mat{D} \mat{H}^{(k)}}_F^2 
    + \frac{b}{2}\norm{\mat{Z}^{(k)} -\mat{H}^{(k)}}_F^2 \nonumber \\
    &+ \frac{b}{2}\norm{\mat{Y}^{(k)} -\mat{H}^{(k)}}_F^2 
     \label{eq:updateX}\\
   \mat{Z}^{(k+1)} = \underset{\mat{Z}}{\arg\min}\, &\mu \mathcal{J}(\mat{Z}) 
   + \frac{b}{2}\norm{\mat{Z} -\mat{H}^{(k+1)}}_F^2\ . \label{eq:updateZ}\\
\mat{Y}^{(k+1)} =  \underset{\mat{Y}}{\arg\min}\,&\lambda \mathcal{Q}(\mat{Y}) 
   + \frac{b}{2}\norm{\mat{Y} -\mat{H}^{(k+1)}}_F^2\ . \label{eq:updateY}
\end{align}
\end{subequations}

\textbf{Data consistency Module - subproblem (\ref{eq:updateX}):} This subproblem takes the form of a quadratic regularized least squares problem, which can be solved as:
\begin{align}
    \mat{H}^{(k+1)} = (\mat{D}^T \mat{D} + 2b\mat{I})^{-1}(\mat{D}^T\mat{Y} + b\mat{Z}^{(k)} + b\mat{Y}^{(k)}).
    \label{eq:x_solution}
\end{align}

\textbf{Denoising Module - subproblem (\ref{eq:updateZ}):} 
The denoising module refines the range image by processing the output received from the data consistency module (as shown in equation \ref{eq:x_solution}). This refinement step, which corresponds to equation (\ref{eq:updateZ}), can be implemented using a deep learning-based denoising network that functions as follows:
\begin{equation} \mat{Z}^{(k+1)} = \mathcal{G}_\theta(\mat{H}^{(k+1)}), \label{eq:denoiser_z} 
\end{equation}
where $\mathcal{G}_\theta$ denotes the denoising neural network that learns to identify and preserve the essential characteristics of range images, effectively serving as a learned prior. The network architecture follows a U-shaped autoencoder design, where the encoder uses 2D convolutional layers to downsample the input, while the decoder employs deconvolutional layers to perform upsampling.
However, since it relies on 2D convolutions, it does not inherently preserve 3D geometric consistency, leading to outliers, particularly in regions with depth discontinuities, sharp edges, or occlusions. These artifacts manifest as spurious points and boundary distortions, which negatively impact SLAM performance.
To mitigate this issue, we introduce an outlier removal module, described in the following, which ensures the structural integrity of the estimated range image while maintaining real-time performance.

\textbf{Outlier Removal - subproblem (\ref{eq:updateY}):} 
This module eliminates outliers introduced by the denoising process, refining the estimated range image. The solution corresponds to the proximal operator of the regularizer $\mathcal{Q}$. Instead of using a neural network, we adopt a cluster-based algorithm \cite{bogoslavskyiFastRangeImagebased2016} that operates without learnable parameters. The algorithm processes the structured 2D range image using breadth-first search (BFS) to label connected components. It scans the image sequentially, starting from the top-left corner, and initiates BFS for each unlabeled pixel. A neighboring pixel is added to the BFS queue based on a criterion involving range measurements $r_1$ and $r_2$ at points $p_1$ and $p_2$, measured by a sensor at $s$. The connectivity angle $\alpha_2$ is:
\begin{equation}
    \alpha_2 = \arctan \left( \frac{\norm{sp_2} \sin \alpha_1}{\norm{sp_1} - \norm{sp_2} \cos \alpha_1} \right),
\end{equation}
where $\alpha_1$ is the known angle between laser beams. A threshold $\theta$ determines connectivity: if $\alpha_2 < \theta$, the points are in separate clusters due to a significant depth change; otherwise, they belong to the same object. This heuristic yet effective method efficiently segments range images, improving SLAM performance.
\textbf{Final iterative solutions}:
Hence, the HQS solver  consists of three interpretable modules that are  the data consistency  solution for estimating the high-resolution range image (\ref{eq:x_up}), the denoising step in equation (\ref{eq:z_up}) and the outlier removal submodule(\ref{eq:y_up}).

\small
\begin{subequations}
\begin{align}
\mat{{H}}^{(k+1)} &= (\mat{D}^T \mat{D} + 2b\mat{I})^{-1}(\mat{D}^T\mat{\mat{Y}} + b(\mat{{Z}}^{(k)} +\mat{{Y}}^{(k)})) \label{eq:x_up} \\
\mat{Z}^{(k+1)} &= \mathcal{G}_\theta(\mat{H}^{(k+1)}) \label{eq:z_up} \\
\mat{Y}^{(k+1)} &= \mathcal{Q}_\theta(\mat{H}^{(k+1)}) \label{eq:y_up}
\end{align}
\label{eq:HQS_final}
\end{subequations}
\normalsize
\textbf{Model-based SR network}:
The learnable components within the denoiser enable us to transform the context-aware optimization problem into an efficient and interpretable deep learning architecture based on established mathematical principles. We accomplish this by implementing the Deep Unrolling framework. Instead of executing numerous iterations of the HQS solver from equation (\ref{eq:HQS_final}), we unroll a limited number K of iterations, with each iteration functioning as a distinct layer in the resulting deep network. This creates a K-layer neural network where each layer maps directly to one HQS iteration, creating an optimal balance between computational efficiency, and model interpretability.
\subsection{End-to-End Architecture: SR-LeGO-LOAM}
  




Having designed the proposed model-based Super-Resolution network, we now introduce the complete end-to-end architecture, as illustrated in Figure \ref{fig:arghitecture}. The proposed architecture extends the LeGO-LOAM framework by integrating the proposed SR model. The objective of this implementation is to incorporate SR functionality while maintaining the real-time performance of the algorithm. The pipeline begins with a sparse point cloud generated by a low resolution LiDAR sensor, which is transformed into a low-resolution range image. The 2D range image is then upscaled using the SR module, that consists of two denoising processes i.e., the autoencoder denoiser (10b) that aims to refines the output of the data-consistency module (1oa) and the outlier removal module  to filter spurious points (10c). The remaining steps of the pipeline follow the standard LeGO-LOAM SLAM process. 
\section{Numerical results}
\subsection{Simulation setup}
We utilized the Ouster LiDAR dataset, which captures a 15-minute drive through San Francisco using an OS-1-64 3D LiDAR sensor. The 64-channel LiDAR point clouds were projected into high-resolution range images of size \(64 \times 1024\) as ground truth, while low-resolution images were generated by extracting 16 out of the 64 channels. During training, we used the AdamW optimizer with PyTorch's default settings and unrolled the HQS solver for \( k = 5 \) iterations, forming a 5-layer deep learning architecture. The model was implemented in C++ and integrated into the LeGO-LOAM pipeline as a preprocessing step, consistently operating in the 2D image domain. It was tested on an RTX 2080 Ti GPU.
\subsection{Evaluation study}
In this section, we present the evaluation results of the proposed SR module with Outlier Removal using the Ouster dataset, which consists of two sequences: a complex 8,000-scan sequence and a smaller 4,000-scan sequence, each subsampled to 16 channels. We compare our method (DU-OR) against state-of-the-art SR LiDAR techniques, including the Super-Resolution Autoencoder (SRAE) \cite{lidar_method}, a simple Deep Unrolling (Simple DU) SR method \cite{my_icip}, and a Transformer-based (VIT) approach \cite{yang2024tuliptransformerupsamplinglidar}. These methods take the downsampled Ouster sequences as input and generate high-resolution 64-channel scans, acting as a preprocessing step for upscaling low-resolution LiDAR data, which are then used as input to the LeGO-LOAM SLAM pipeline. Additionally, we also evaluate LeGO-LOAM performance by directly feeding the low-resolution scans (LiDAR-16) without any processing. As a metric, we use the Absolute Pose Error (APE), with the ground truth being the LeGO-LOAM output using the original 64-channel LiDAR. We also report the execution time for each SR technique and the network size in terms of the number of parameters.
\begin{table}
  \centering
  \small
  \setlength{\tabcolsep}{4pt} 
  \caption{Absolute Pose Error (RMSE) w.r.t. translation part (m), FPS, and Number of Parameters.}
  \resizebox{0.4\textwidth}{!}{
  \begin{tabular}{cccc}
    \toprule
    Method & RMSE [m]  & FPS & Parameters \\
    \midrule
    LiDAR-16  & 11.58 / 6717  & - & - \\
    SRAE \cite{lidar_method}  & 5.53 / 67.35  & 5 & 35M \\
    Simple DU \cite{my_icip}  & 3.78 / 31.71  & 5 & 0.1M \\
    VIT \cite{yang2024tuliptransformerupsamplinglidar}  & 3.58 / 28.64  & 1 & 50M \\
    DU-OR (ours)  & \textbf{2.35} / \textbf{22.64}  & \textbf{400} & 0.2M \\
    \bottomrule
  \end{tabular}}
    \label{tab:lego_error}
\end{table}
Table \ref{tab:lego_error} presents the RMSE values for both sequences in a single column, with the results for the 4,000-scan sequence listed first, followed by those for the 8,000-scan sequence, separated by a slash. The results highlight the limitations of the 16-channel LiDAR, with an APE RMSE of 11.58m for 4,000 scans, which sharply increases to 6,717m for 8,000 scans, demonstrating the challenges of low-resolution data in complex environments due to the sparsity of point clouds. The proposed DU-OR method consistently outperforms all tested SR approaches, achieving APE RMSE reductions ranging from 34\% to 58\% for 4,000 scans and 20\% to 66\% for 8,000 scans. Finally, Figure 2 validates the above mentioned results. 
Beyond accuracy, DU-OR excels in real-time performance, achieving 400 fps, compared to 5 fps for SR-AE and 1 fps for VIT, while containing $\textbf{99\%}$ less parameters. The slower speeds of SR-AE \cite{lidar_method} and VIT methods \cite{yang2024tuliptransformerupsamplinglidar}  result from Monte Carlo-based inference, requiring multiple evaluations per point cloud. In contrast, DU efficiently captures LiDAR structures, reconstructing high-resolution point clouds with a single inference, making it ideal for real-time autonomous applications. 
\begin{figure}
 \begin{subfigure}{0.23\textwidth}
     \includegraphics[width=\textwidth]{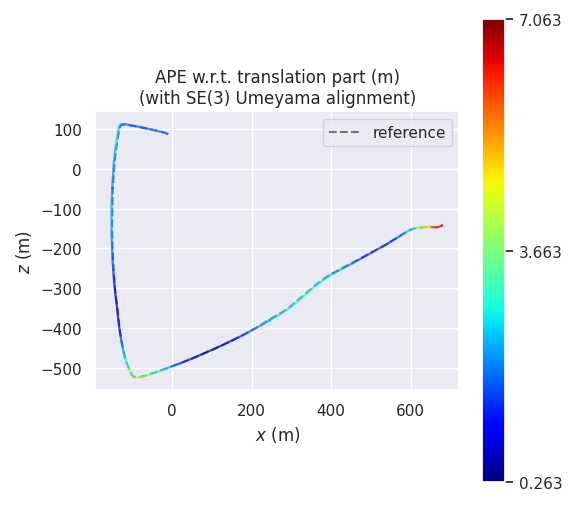}
     \caption{proposed DU heatmap}
     \label{fig:myDU_6000}
 \end{subfigure}
 \hfill
 \begin{subfigure}{0.23\textwidth}
     \includegraphics[width=\textwidth]{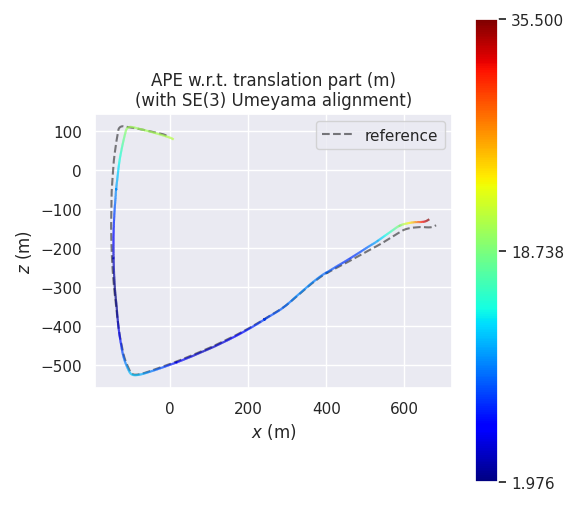}
     \caption{LiDAR-16 heatmap}
     \label{fig:icip_6000}
 \end{subfigure}
 \medskip
 \begin{subfigure}{0.23\textwidth}
     \includegraphics[width=\textwidth]{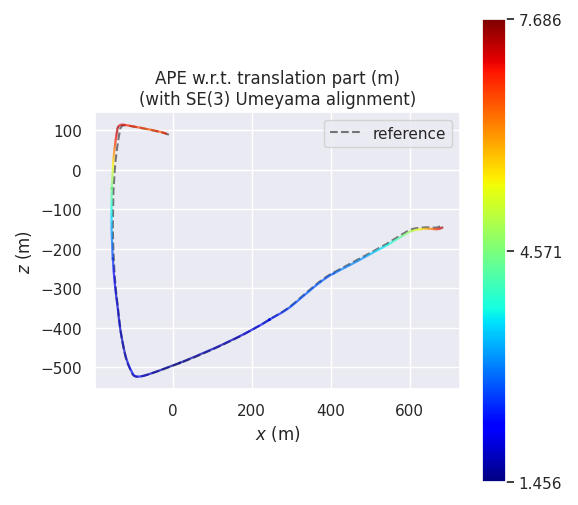}
     \caption{VIT \cite{yang2024tuliptransformerupsamplinglidar}  heatmap }
     \label{fig:sr-ae-6000}
 \end{subfigure}
 \caption{Heatmaps for the proposed DU, the 16-channel LiDAR and the Transformer based method \cite{yang2024tuliptransformerupsamplinglidar} using as reference trajectory the path derived from the LiDAR-64.  }
 \label{fig:l_6000}
\end{figure}
\section{Conclusion}
This work presents a DU-based SR model with an integrated outlier removal module to enhance the accuracy and efficiency of low-resolution LiDAR sensors for SLAM applications. By leveraging a model-based optimization approach, our method reconstructs high-resolution point clouds while maintaining real-time performance. The proposed SR model was integrated and evaluated within a state-of-the-art LiDAR SLAM framework, demonstrating significant improvements in pose estimation accuracy and efficiency over existing SR methods. These findings highlight the effectiveness of outlier-aware Super-Resolution in improving SLAM performance and its potential extension to other research areas such as object detection.


\label{sec:refs}
\small
\bibliographystyle{IEEEbib}

\bibliography{mybibfile}

\end{document}